\documentclass[conference]{IEEEtran}
\IEEEoverridecommandlockouts
\usepackage{amsmath,amsfonts, amssymb}
\usepackage{array}
\usepackage[caption=false,font=normalsize,labelfont=sf,textfont=sf]{subfig}
\usepackage{textcomp}
\usepackage{stfloats}
\usepackage{url}
\usepackage{verbatim}
\usepackage{booktabs}
\usepackage{graphicx}
\usepackage{algorithm}
\usepackage{xcolor}
\usepackage[noend]{algorithmic} 
\def\BibTeX{{\rm B\kern-.05em{\sc i\kern-.025em b}\kern-.08em
    T\kern-.1667em\lower.7ex\hbox{E}\kern-.125emX}}
\usepackage{balance}
\newcommand{\method}{MASK} 

\begin{document} 
\title{MASK: Multi-Agent Semantic K-Scheduling for Risk-Sensitive 6G Robotics} 

\author{Ahmet Günhan Aydın, Elif Tugce Ceran
 \thanks{Authors are with the Department of Electrical and Electronics Engineering, Middle East Technical University, Ankara, 06800, Turkey, e-mail: gunhan.aydin@metu.edu.tr, elifce@metu.edu.tr. A. G. Aydın is also with Aselsan Inc., Ankara, Turkey, e-mail: gunaydin@aselsan.com.}}

\markboth{Journal of \LaTeX\ Class Files,~Vol.~18, No.~9, September~2020}%
{How to Use the IEEEtran \LaTeX \ Templates}

\maketitle

\begin{abstract}
Realizing the vision of 6G connected robotics requires reconciling high-performance collaborative control with the rigid spectral limitations of physical wireless channels. In realistic collaborative sensing scenarios, spectral resources are quantized into finite physical resource blocks or orthogonal subcarriers, rendering simultaneous transmission by all agents infeasible. To address this, we propose Multi-Agent Semantic K-Scheduling (\method{}), a control architecture designed to sustain robust, risk-aware coordination under strict instantaneous bandwidth caps. We introduce \textit{Arbiter-Assisted Semantic Information Gating} (A-SIG), a lightweight coordination mechanism that enforces hard access constraints by scheduling only the top-$K$ agents based on locally computed semantic importance scores. By aggregating these prioritized observations into a compact latent state, a self-supervised global encoder enables a distributional policy to mitigate tail risks despite data sparsity. We evaluate \method{} across diverse benchmarks, demonstrating that it matches the performance of communication-unconstrained baselines even when channel access is restricted to a small fraction of the swarm size. Furthermore, the framework exhibits inherent resilience to packet erasures, validating semantic scheduling as a critical enabler for resource-constrained 6G systems.
\end{abstract}


\section{Introduction}

\IEEEPARstart{T}{he} deployment of autonomous robotic swarms in next-generation 6G networks necessitates a paradigm shift in Multi-Agent Reinforcement Learning (MARL) \cite{marl_abbrv}. Whether in non-terrestrial networks (NTN) \cite{Ji2023,Lyu2024} or industrial IoT, the fundamental challenge is coordinating decentralized agents under strict physical layer constraints. While agents must exchange local observations to resolve partial observability \cite{marl_partial_obs}, the underlying 6G infrastructure cannot physically support simultaneous broadcasting by all nodes. In realistic edge scenarios, time-frequency resources are quantized into limited physical resource blocks (PRBs) or orthogonal subcarriers, imposing a hard instantaneous channel access constraint that limits the number of concurrent transmissions to at most $K$. Existing control frameworks, however, rarely account for these sharp physical limits while simultaneously ensuring safety in stochastic environments.

Prior research has generally approached coordination through two disjoint lenses. The first focuses on \textit{information aggregation}. Architectures like CommNet \cite{sukhbaatar2016commnet}, TarMAC \cite{das2019tarmac}, and MASIA \cite{chen2022masia} aim to reconstruct global states but rely on communication-intensive protocols that assume ideal, high-capacity channels. Furthermore, these methods largely optimize for risk-neutral expected returns, ignoring the catastrophic tail risks inherent in dynamic physical environments.

The second lens focuses on \textit{risk-sensitivity}. To operate safely in unpredictable domains like autonomous driving, agents must account for rare, high-cost tail events. Distributional RL \cite{bellemare2017c51, dabney2018qr} and frameworks like RiskQ \cite{riskq_placeholder} address this by explicitly modeling the full return distribution rather than optimizing a simple mean. These algorithms further apply risk functionals to the return distributions, such as Conditional Value-at-Risk (CVaR)~\cite{ROCKAFELLARcvar}, to emphasize adverse tail outcomes. However, these risk-aware policies are typically trained on purely local views. This creates a dangerous blind spot: an agent may behave conservatively based on its own sensor data while remaining oblivious to a critical threat visible only to a distant teammate.

To bridge this disconnect, we propose a joint control-communication architecture designed explicitly for resource-constrained 6G interfaces. We argue that agents must learn not only \emph{how} to act safely but \emph{which} data is semantically essential to transmit when channel access is competitively arbitrated. This reflects the 6G vision of \textit{semantic communication}, optimizing data exchange for task utility rather than throughput \cite{Gunduz2023,Uysal2022}. We introduce Multi-Agent Semantic K-Scheduling (\method{}), a framework that replaces unstructured broadcasting with \textit{Arbiter-Assisted Semantic Information Gating} (A-SIG). In this scheme, a Central Arbiter strictly enforces physical channel access by granting transmission rights to the top-$K$ agents with the highest semantic importance scores, ensuring robust global coordination without violating physical channel constraints.

Our main contributions are:
\begin{itemize}
    \item We propose \textit{Arbiter-Assisted Semantic Information Gating} (A-SIG), a differentiable scheduling mechanism in which a Central Arbiter enforces hard physical channel access constraints (top-$K$) by ranking locally computed semantic importance scores.
    \item We introduce \method{}, a unified architecture combining A-SIG with a self-supervised global encoder and a risk-sensitive distributional policy, optimizing joint safety and performance under strict channel access constraints.
    \item We empirically demonstrate that \method{} matches the performance of unconstrained baselines even when channel access is restricted to a small fraction of the swarm size. Furthermore, our results validate that the framework maintains robust performance under random packet erasures, making it suitable for realistic, unreliable 6G robotic networks.
\end{itemize}

\section{Related Work}
\label{sec:related}

Coordination under partial observability is a cornerstone of networked robotics. Early differentiable communication works like CommNet \cite{sukhbaatar2016commnet} and DIAL \cite{Foerster2016} allowed gradient propagation across channels, while attentional models (ATOC \cite{jiang2018atoc}, TarMAC \cite{das2019tarmac}) improved scalability. Recent transformer-based approaches (e.g., MACTAS \cite{mactas_transformer}) further enhance state aggregation but rely on continuous, high-bandwidth exchange.

For 6G edge applications, however, bandwidth is limited. Approaches like IC3Net \cite{singh2018ic3net} and NDQ \cite{wang2020ndq} attempt to gate communication, while Tung et al. \cite{Tung2021} address noisy channels. While MASIA \cite{chen2022masia} effectively constructs global representations, it remains communication-intensive. In contrast, our A-SIG mechanism embraces the \textit{semantic communication} philosophy: we do not aim to reconstruct the full bit-stream of observations, but rather to transmit only the semantically important observations necessary for the joint risk-sensitive control task. While \cite{su2025goal} selects specific semantic features per agent, \method{} schedules the top-$K$ agents for risk-sensitive control.

Standard MARL optimizes expected returns, ignoring the variance inherent in dynamic environments \cite{rashid2018qmix}. Distributional RL \cite{bellemare2017c51} overcomes this by modeling the return distribution, enabling risk measures like CVaR \cite{ROCKAFELLARcvar}. In multi-agent settings, RiskQ \cite{riskq_placeholder} applies distributional factorization to manage collective risk. However, RiskQ lacks a communication mechanism, forcing agents to make risk assessments based solely on local views. \method{} unifies these domains, using A-SIG to resolve information uncertainty, thereby providing the global context necessary for valid risk-sensitive decision-making.

\section{Methodology: Multi-Agent Semantic K-Scheduling (\method{})}
\label{sec:method}

We formulate the cooperative multi-agent task as a Decentralized Partially Observable Markov Decision Process (Dec-POMDP), defined by the tuple $\mathcal{G} = \langle \mathcal{S}, \mathcal{A}, P, R, \Omega, O, N, \gamma, \mathcal{M} \rangle$. At each timestep $t$, each agent $i$ receives a local observation $o_i^t \sim O(s, i)$ from the observation space $\Omega$ along with a message $m_i^t \in \mathcal{M}$ from other agents. Based on these inputs, the agent selects an environmental action $a_i^t \in \mathcal{A}$ and a binary communication action $b_i^t \in \{0, 1\}$. 
Here, $b_i^t=1$ denotes transmitting the observation, while $b_i^t=0$ denotes silence. The joint action $A^t$ drives the state transition $s^{t+1} \sim P(s^t, A^t)$ and reward $r^t = R(s^t, A^t)$. Unlike prior works that optimize the expected return $\mathbb{E}[\sum \gamma^t r^t]$ under idealized communication, we address a setting where channel access is explicitly constrained and agents must be robust to outcome uncertainty.

Our objective is to maximize a \textit{risk-sensitive} functional $\rho$ of each agent’s return distribution $Z_i$, defined over the discounted return $\sum_{k=0}^{T} \gamma^k r^{t+k}$. We adopt the Distortion Risk Measure (DRM) framework~\cite{drm_hardy}, and in particular optimize the Conditional Value-at-Risk~\cite{ROCKAFELLARcvar} to ensure robustness to adverse tail outcomes. The resulting objective is

\begin{equation}
    \rho_{\text{CVaR}}(Z_i) = \frac{1}{\alpha} \int_0^\alpha F_{Z_i}^{-1}(\omega) \, d\omega,
\end{equation}
where $\alpha \in (0, 1]$ is the risk tolerance and $F_{Z_i}^{-1}$ is the quantile function.

\begin{figure*}[htbp]
  \centering
  \includegraphics[width=0.9\linewidth]{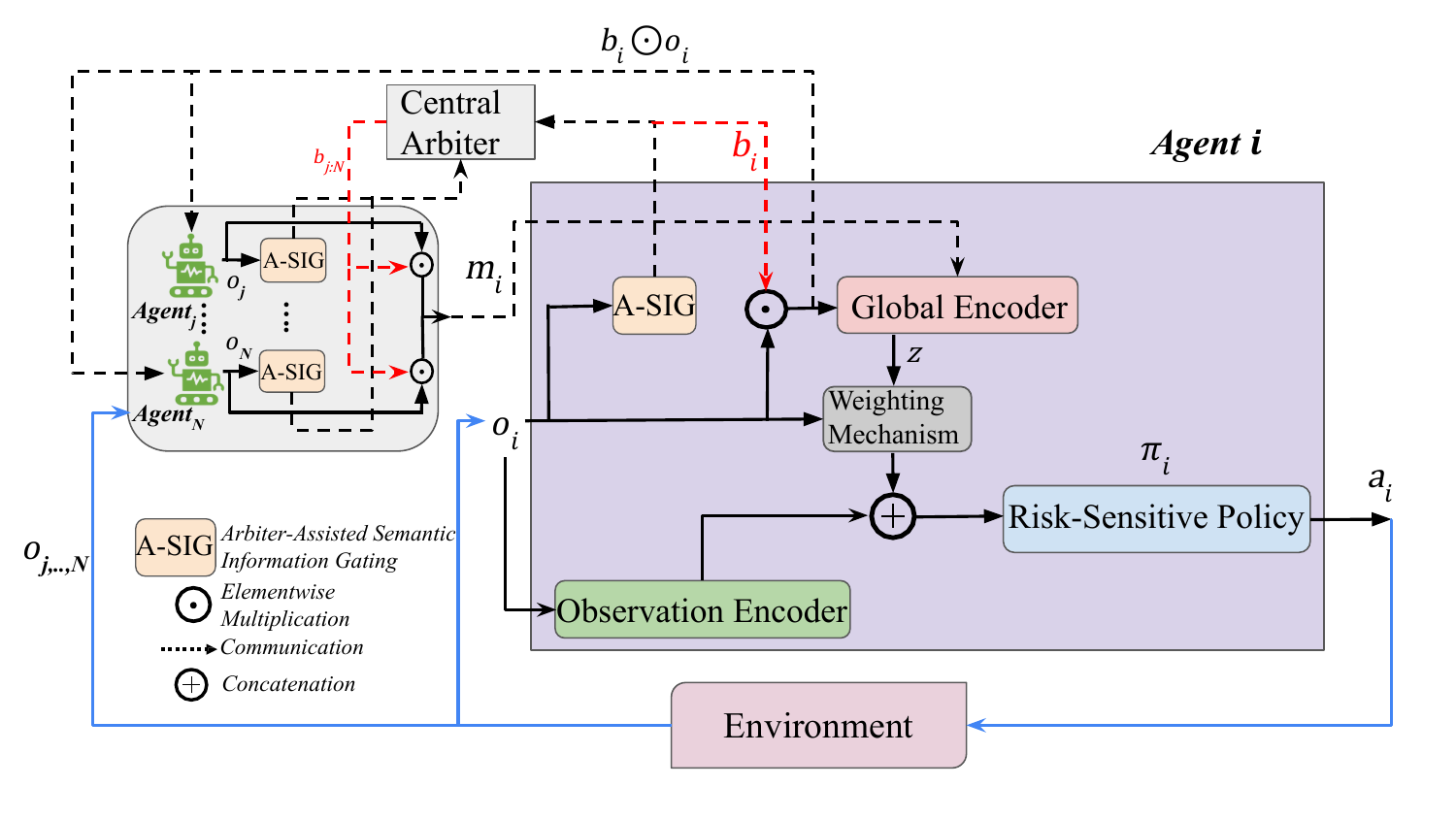}

  \caption{The \method{} agent and the system architecture. Agents employ an A-SIG module to selectively broadcast critical observations with the assistance of a Central Arbiter. For each agent $i$, the received messages $m_i$ are processed by a global encoder to reconstruct a latent global state $z$, which is combined with the encoded local observation $o_i$ to produce a risk-sensitive action $a_i$. Solid arrows denote local computation; dashed arrows represent communication flows. Red dashed lines indicate the gating feedback (transmission decision) from the Central Arbiter. Element-wise gating filters the observations to regulate bandwidth usage during both centralized training and decentralized execution.}
  \label{fig:arch-method}
\end{figure*}

To solve this under channel access constraint, we propose \method{}, a framework following the Centralized Training with Decentralized Execution (CTDE) paradigm (see Fig.~\ref{fig:arch-method}). The architecture unifies three key modules:
(1) an A-SIG module that computes local importance scores to arbitrate channel access via a Central Arbiter;
(2) a self-supervised per-agent global encoder that aggregates the sparse, gated observations into a latent global state $z^t$;
(3) a \textit{Distributional Risk-Sensitive Policy} that combines $z^t$ with local views to estimate the return distribution $Z_i$. We classify this execution model as decentralized because the Central Arbiter's role is limited to low-bandwidth control signaling (scheduling), ensuring that relatively high-bandwidth semantic data transmission remains a peer-to-peer, agent-driven process.

During centralized training, we minimize a joint loss $L_{total} = L_{\text{RL}} + \lambda_{\text{repr}} L_{\text{repr}}$ that simultaneously optimizes the distributional value factorization, the global state reconstruction, while providing communication efficiency through the A-SIG mechanism. During both training and execution, the A-SIG module enables a Central Arbiter to enforce strict channel access caps by granting transmission rights only to agents with the highest importance scores, ensuring that the latent representation $z^t$ is constructed from the most semantically relevant information. 

\subsection{Arbiter-Assisted Semantic Information Gating (A-SIG)}
\label{subsec:esig}

To address communication resource constraints while maintaining global coordination, we introduce the \textit{Arbiter-Assisted Semantic Information Gating} module. Unlike fully decentralized gating mechanisms where agents make isolated transmission decisions, A-SIG employs a feedback-driven architecture that decouples low-bandwidth control signaling from high-bandwidth data transmission.

The gating process operates in three distinct phases:

\begin{enumerate}
    \item \textbf{Local Importance Quantification:} Each agent $i$ is equipped with a lightweight communication scorer network, denoted by $C_{\omega}$, parameterized as a 2-layer MLP. At each timestep $t$, the agent processes its local observation $o_i^t$ to generate a scalar importance score $c_i^t = C_{\omega}(o_i^t) \in \mathbb{R}$. This score represents the agent's estimated utility of broadcasting its current state to the team.
    
    \item \textbf{Global Arbitration at the Central Arbiter:} The importance scores $\mathbf{c}^t = \{c_1^t, \dots, c_N^t\}$ are transmitted to the Central Arbiter over a low-rate control channel.
    Compared to full observation vectors, this scalar transmission incurs negligible bandwidth overhead, despite scaling with swarm size. 
    The Central Arbiter aggregates the scores and performs a global ranking operation to identify the subset of agents with the most critical information. A binary transmission mask $\mathbf{b}^t \in \{0, 1\}^N$ is generated based on a system-wide channel access constraint $K$ (e.g., selecting the top-$K$ highest scores) via the indicator function $b_i^t = \mathbb{I}(c_i^t \geq \text{Rank}_K(\mathbf{c}^t))$, where $b_i^t=1$ indicates permission to transmit.
    
    \item \textbf{Gated Data Transmission:} The binary decision $b_i^t$ is sent back to agent $i$ as immediate feedback over the control channel. Agents with $b_i^t=1$ broadcast their observations to the multi-agent network, while others remain silent, significantly reducing channel congestion. Each agent $i$ receives the sparse message set $m_i^t = \{b_j^t \cdot o_j^t\}_{j \neq i}$.
\end{enumerate}

To enable end-to-end training of the scorer network $C_{\omega}$ despite the non-differentiable nature of the ranking and thresholding operation, we employ a Straight-Through Estimator (STE)~\cite{bengio2013ste}. During the forward pass, the hard binary mask $\mathbf{b}^t$ is applied strictly to gate the information flow. However, during the backward pass, the STE allows gradients to bypass the discrete thresholding, treating the decision as a continuous function of the scores. This proxy gradient allows the scorer network to learn to assign higher scalar values to observations that yield lower global loss, effectively aligning local importance estimation with the team's cooperative objective.

\subsection{Self-Supervised Selective State Aggregation}
The global encoder $E_\phi$ processes the sparsely filtered observations $O^t_{filt} = \{b_i^t \cdot o_i^t\}_{i=1}^N$ (where non-transmitting agents contribute zero) to produce a compact latent global state $z^t = E_\phi(O^t_{filt})$. This process is consistent across CTDE, where agents reconstruct $z^t$ from received observations $O^t_{filt}$.
$E_\phi$ employs a self-attention mechanism to capture agent interactions.
The filtered observations are projected into query, key, and value spaces via $\{Q,K,V\} = \text{MLP}_{\{Q,K,V\}}(O^t_{filt})$ to compute the hidden state matrix $M = \text{softmax}(Q K^\top / \sqrt{d_k})V$, where $d_k$ is the key dimension.
The aggregated latent vector $z^t$ is derived from $M$ via an integration network. To ensure $z^t$ is globally informative and temporally predictive, we optimize a joint self-supervised objective $L_{\text{repr}} = L_{\text{AE}} + \lambda_{\text{pred}} L_{\text{pred}}$.
The reconstruction loss $L_{\text{AE}}$ trains a decoder $D_\zeta$ to recover $s^t$, while the multi-step prediction loss $L_{\text{pred}}$ trains a transition model $L_\xi$ to predict future latents  over a horizon $H$:
\begin{align}
    L_{\text{AE}} &= \mathbb{E}\big[\| D_\zeta(z^t) - s^t \|^2\big], \\ \quad
L_{\text{pred}} &= \mathbb{E}\left[\sum_{k=1}^{H} \| L_\xi(\hat{z}^{t+k-1}, A^{t+k-1}) - z^{t+k} \|^2\right].
\end{align}
This objective ensures the encoder captures essential global dynamics despite sparse inputs.

\subsection{Risk-Sensitive Policy Head}
\label{sec:risk_policy_head}
To address outcome uncertainty, the agent integrates its local embedding $f_o(o_i^t)$ with the global context $z^t$ (modulated by a learned weighting vector $g_i^t$) to form the policy input $x_i^t = \text{concat}(f_o(o_i^t),\, g_i^t \odot z^t,\, \textit{id}_i)$. This input updates the recurrent hidden state $h_i^t$.
In practice, the full action--observation history $\tau_i$ is summarized by the recurrent hidden state $h_i^t$.
We adopt the Implicit Quantile Network (IQN) framework \cite{dabney2018iqn}.  To approximate the return distribution, quantile fractions $\omega \sim U(0,1)$ are mapped to an embedding space via $\phi(\omega)$ (e.g., cosine embedding) and combined with $h_i^t$ to generate the per-action quantiles $\theta_i(\tau_i, a_i, \omega)$.
During centralized training, a monotonic distributional mixer aggregates these into a joint distribution $\theta_{\text{tot}}$ using weights $k_i \ge 0$ to ensure decomposability:
\begin{equation}
\theta_{\text{tot}}(\tau, a, \omega) = \sum_{i=1}^N k_i(\tau, s, \omega) \, \theta_i(\tau_i, a_i, \omega).
\label{eq:q_tot}
\end{equation}
The network is trained via quantile regression minimizing the quantile Huber loss $\mathcal{L}_\kappa$ against a target $y^t$ constructed using a risk-sensitive Bellman operator that selects actions maximizing the risk measure $\rho$ on the next-state distribution $Z(\tau^{t+1}, a')$. Here, $Z$ implicitly represents the joint distribution of all agents (omitting indices for clarity), and the prime notation ($Z'$, $\beta'$) denotes the usage of the target network. The objective is defined as:
\begin{align}
L_{\text{RL}} &= \mathbb{E}[\mathcal{L}_\kappa(y^t - \theta_{\text{tot}})], \\\quad y^t &= r^t + \gamma \theta_{\text{tot}}(\tau^{t+1}, \arg \max_{a'} \rho(Z'), \omega; \beta').
\label{eq:rl_loss}
\end{align}
The final joint objective updates all components: $L_{\text{total}} = L_{\text{RL}} + \lambda_{\text{repr}} L_{\text{repr}}$. The centralized training procedure is summarized in Algorithm~\ref{alg:main}. During decentralized execution, each agent independently reconstructs the latent state from received messages using its local copy of the global encoder.

\begin{algorithm}[ht]
\small
\caption{\method{} Training Algorithm}
\label{alg:main}
\begin{algorithmic}[1]
\STATE Initialize agent networks $\{Q_i\}_{i=1}^N$ with encoder $E_\phi$ and A-SIG $C_\omega$, latent model $L_\xi$, mixer $Q_{\text{mix}}$
\STATE Initialize target networks $\{Q'_{i}\}_{i=1}^N$, $Q'_{\text{mix}}$, replay buffer $\mathcal{D}$, and max episodes $M$

\FOR{episode $= 1$ to $M$}
    \STATE Reset environment and observe global state $s^1$ and observations $O^1 = \{o_1^1, \ldots, o_N^1\}$ 
    \STATE Initialize trajectory buffer $\tau \leftarrow \emptyset$
    \FOR{timestep $t = 1$ to $T$}
    \STATE Compute importance scores $c_i^t = C_\omega(o_i^t), i=1 \dots N$
    \STATE Transmit scores $\mathbf{c}^t$ to the Central Arbiter
    \STATE Central Arbiter computes rank threshold: $\delta^t = \text{Rank}_K(\mathbf{c}^t)$
    \STATE Generate transmission mask $b_i^t = \mathbb{I}(c_i^t \ge \delta^t), i=1 \dots N$
    \STATE Filter observations $O^t_{\text{filt}} = \{b_i^t \cdot o_i^t\}_{i=1}^N$
    \STATE Compute latent state $z^t = E_\phi(O^t_{\text{filt}})$
        \FOR{agent $i = 1$ to $N$}
            \STATE Form input $x_i^t = [f_o(o_i^t),\, g_i^t \odot z^t, \textit{id}_i]$
            \STATE Compute quantile distribution $Z_i(x_i^t, \cdot, a)$ via $Q_i$
            \STATE Compute risk-sensitive values $Q_{i,\rho}(a) = \rho(Z_i(x_i^t, \cdot, a))$
            \STATE Select action $a_i^t = \epsilon\text{-greedy}(Q_{i,\rho})$
        \ENDFOR
        \STATE Execute $A^t = \{a_i^t\}_{i=1}^N$ and observe $r^t$, $O^{t+1}$, $\text{terminated}$
        \STATE Append $(s^t, O^t, A^t, r^t, O^{t+1}, \text{terminated})$ to $\tau$
    \ENDFOR
    \STATE Store $\tau$ in replay buffer $\mathcal{D}$

    \IF{$\mathcal{D}$ contains sufficient samples}
        \STATE Sample minibatch $\mathcal{B}$ from $\mathcal{D}$

        \STATE Compute representation loss as $L_{\text{repr}} = L_{\text{AE}} + \lambda_{\text{pred}} L_{\text{pred}}$

        \STATE Compute joint quantiles $\theta_{\text{tot}}$ using Eq.~\eqref{eq:q_tot} with online networks $\{Q_i\}_{i=1}^N$ and $Q_{\text{mix}}$
        \STATE Compute target quantiles $y^k$ (with target networks $\{Q'_{i}\}_{i=1}^N$, $Q'_{\text{mix}}$) and $L_{\text{RL}}$ using Eq.~\eqref{eq:rl_loss}

        \STATE Minimize total loss $L_{\text{total}} = L_{\text{RL}} + \lambda_{\text{repr}} L_{\text{repr}}$, update the network parameters
       
        \IF{episode \% \text{target\_update\_interval} $= 0$}
            \STATE Update $\{Q'_{i}\}_{i=1}^N$ and $Q'_{\text{mix}}$
        \ENDIF
    \ENDIF
\ENDFOR
\end{algorithmic}
\end{algorithm}

\section{Experiments}
\label{sec:experiments}

We evaluate \method{} on a suite of benchmarks designed to assess performance under varying partial observability, coordination complexity, and outcome stochasticity.

The \textit{Hallway Group} task~\cite{chen2022masia, wang2020ndq} is a partially observable navigation problem emphasizing temporal coordination. Agents are divided into two groups that must reach a central goal at different, predefined times. This constraint introduces strong coupling, requiring global coordination to resolve symmetry and avoid congestion. This setup proxies 6G-enabled warehouse logistics, where fleets of autonomous robots must coordinate passage through shared bottlenecks. The requirement to stagger arrival times mirrors time-sensitive networking in 6G, where precise scheduling is critical to prevent both physical collisions and spectrum congestion in dense swarms.

The \textit{Multi-Agent Car Following (MACF)} task~\cite{riskq_placeholder} extends a risk-sensitive driving environment~\cite{risk_sensitive_env} to a cooperative setting where two agents control their acceleration to platoon toward a shared goal. Under partial observability, agents receive positive rewards for maintaining formation within a limited sensing range, while simultaneously accruing negative step rewards that incentivize speed. The environment explicitly evaluates risk-sensitive control via stochastic crash dynamics: exceeding specific speed thresholds incurs probabilistic collisions, forcing agents to balance progress efficiency against catastrophic failure. This formulation effectively mimics safety-critical V2V platooning, where connected autonomous vehicles must optimize traffic flow while strictly adhering to safety margins under mechanical or sensing uncertainties.

To isolate the contributions of semantic gating and risk sensitivity, we compare \method{} against baselines representing distinct combinations of information handling and objective formulations.
QMIX \cite{rashid2018qmix} serves as the foundational risk-neutral baseline, employing monotonic value factorization on local observations to optimize expected return.
RiskQ \cite{riskq_placeholder} extends QMIX to distributional RL, enabling risk-sensitive policies via quantile regression, but remains limited to local observations.
MASIA \cite{chen2022masia} addresses partial observability by aggregating a global state from all agent observations to inform local policies, yet remains risk-neutral.
\method{} (Ours) unifies these approaches, integrating self-supervised global state aggregation with a risk-sensitive distributional policy, while uniquely employing A-SIG to optimize communication efficiency against hard channel access constraints.

\subsection{Simulation Results}

We benchmark \method{} against the baselines to evaluate its efficacy across the selected tasks. The results empirically validate that jointly addressing information and outcome uncertainty under strict communication constraints allows for robust coordination, demonstrating the core advantages of our architecture. All learning curves report the mean and standard deviation over at least three independent runs. Models are evaluated every 10k training steps, averaging results over 10 episodes for MACF and 100 episodes for Hallway Group.

\begin{figure}[h]
    \centering
    \includegraphics[width=\linewidth]{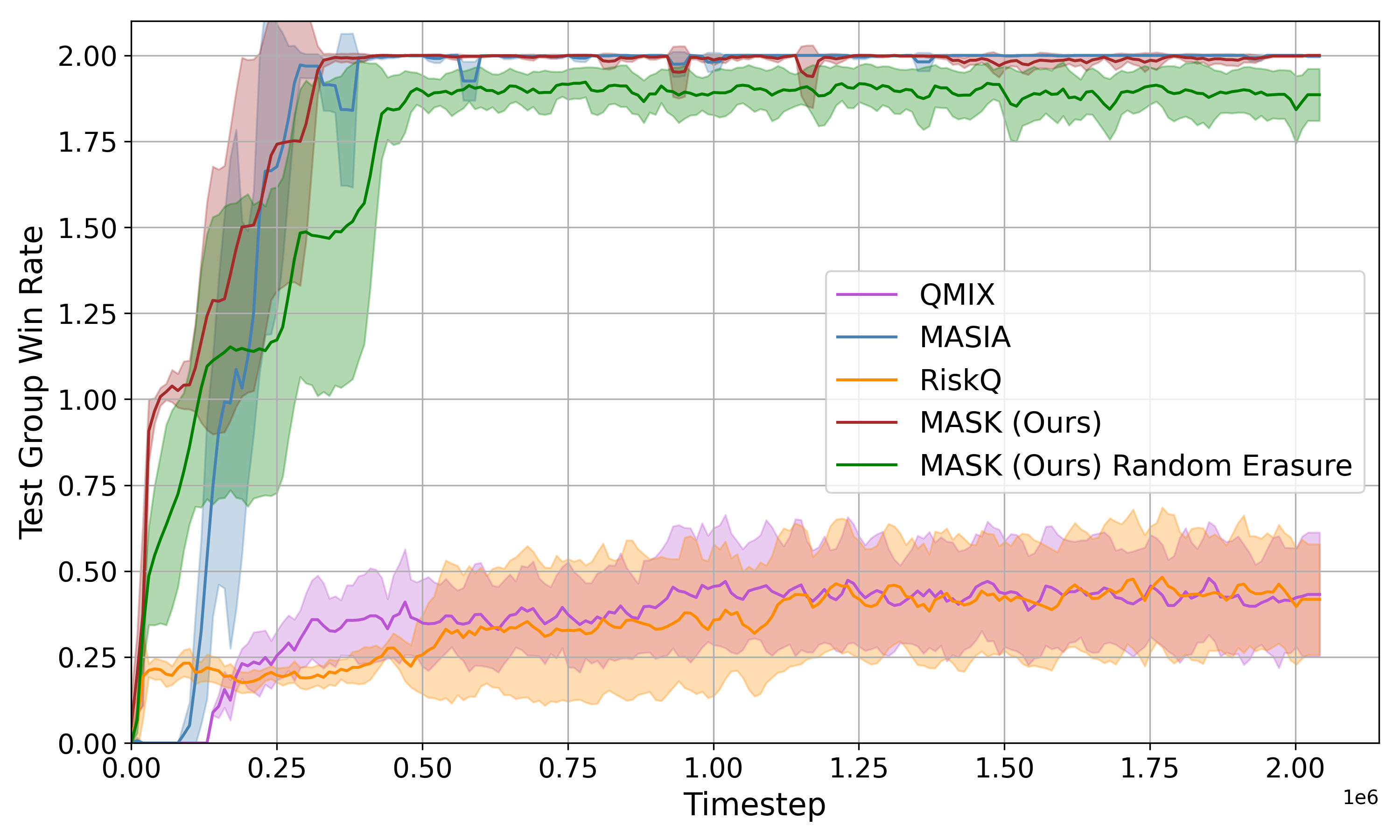}
    \caption{Test group win rates over training timesteps in the Hallway Group environment with $N=5$ agents divided into two groups. 
    For \method{}, the communication parameter is set to $K = 4 \;(0.8N)$ to ensure robust and fast convergence in this high-coordination setting.}
    \label{fig:hallway_group}
\end{figure}

We first consider the Hallway Group environment, for which the results are shown in Figure~\ref{fig:hallway_group}, a setting that poses a coordination challenge under partial observability. Agents must precisely time their actions to pass through a narrow bottleneck, making effective coordination essential. Local-observation baselines, such as QMIX and RiskQ, fail to establish stable coordination and achieve group win rates between 0.2 and 0.6. In contrast, \method{} quickly converges to the optimal group win rate of 2.0, indicating that both groups consistently reach the goal. This performance matches that of the full-communication MASIA baseline, yet \method{} achieves it while enforcing a hard communication constraint of $K=4$ transmitting agents. These results demonstrate that the proposed A-SIG mechanism successfully prioritizes task-relevant information, enabling accurate global state reconstruction.

\begin{figure}[h]
    \centering
    \includegraphics[width=\linewidth]{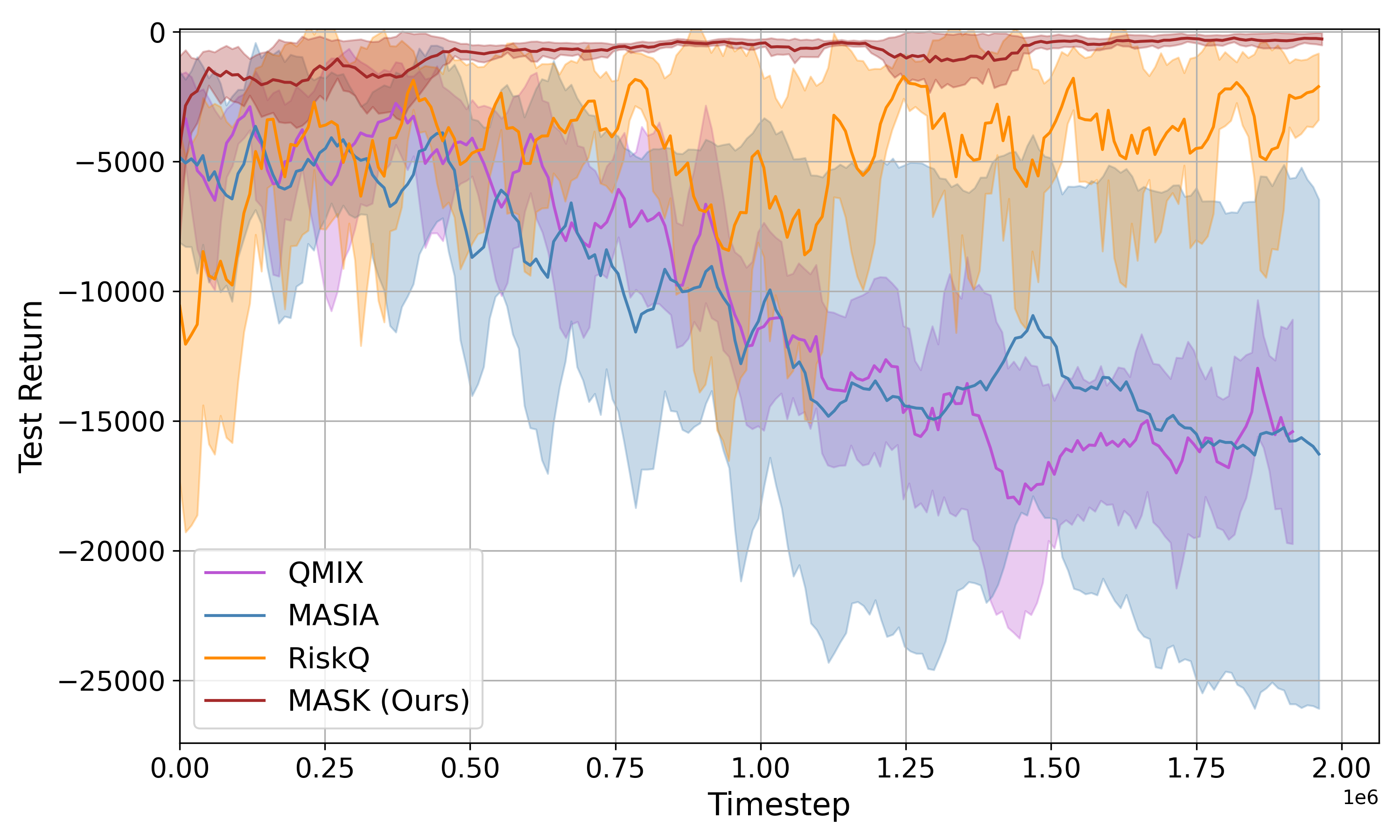}
    \caption{Test returns over training timesteps on MACF with $N=2$ agents. 
    The communication parameter is set to $K = 1 \; (0.5N)$.}
    \label{fig:macf}
\end{figure}

Figure~\ref{fig:macf} presents the results from the MACF environment, a scenario designed to test safety and risk-aware control in the presence of stochastic collision dynamics. In this domain, risk-neutral baselines like MASIA and QMIX display instability and poor performance, as they overlook the tail risks inherent in unsafe maneuvers. Conversely, \method{} demonstrates consistent stability and high returns throughout the training process. Notably, even with the channel access capped at $K=1$, the results indicate that permitting a single agent to transmit per timestep is sufficient to effectively mitigate risk. This confirms that risk-aware decision-making can be sustained with minimal but strategic communication, making the approach highly applicable to practical, bandwidth-limited robotic networks.

To evaluate robustness under realistic 6G edge conditions, we test \method{} in the Hallway Group environment under a random erasure channel where each agent’s importance score reaches the Central Arbiter with probability $0.7$, and is otherwise received as zero. Since importance scores $c_i^t \in \mathbb{R}$ are unbounded, a zero-filled erasure may inadvertently be ranked higher than a valid negative score. We apply Top-$K$ scheduling ($K=4$) based on these received values; however, agents with erased scores are physically unable to transmit observations even if selected by the Central Arbiter. Despite this stochastic mismatch between scheduling and transmission, \method{} attains an average group win rate of $\approx 1.9$ (vs. $2.0$ for perfect channels) while maintaining an effective post-erasure communication rate of $\approx 0.6$, as shown in Figure~\ref{fig:hallway_group}. This demonstrates that the global encoder reliably reconstructs the global state from stochastically incomplete inputs, supporting robust coordination under packet erasures typical of high-mobility and NTN 6G scenarios~\cite{Ji2023}.

In summary, \method{} with A-SIG demonstrates versatile performance across distinct challenges. It acts as a dynamic switch, matching state-of-the-art communication methods in highly partially observable tasks (Hallway Group) and setting new benchmarks for stability in safety-critical environments (MACF) where traditional risk-neutral baselines falter.

\subsection{Impact of Channel Access Constraint ($K$)}

\begin{figure}[h]
    \centering
    \includegraphics[width=0.95\linewidth]{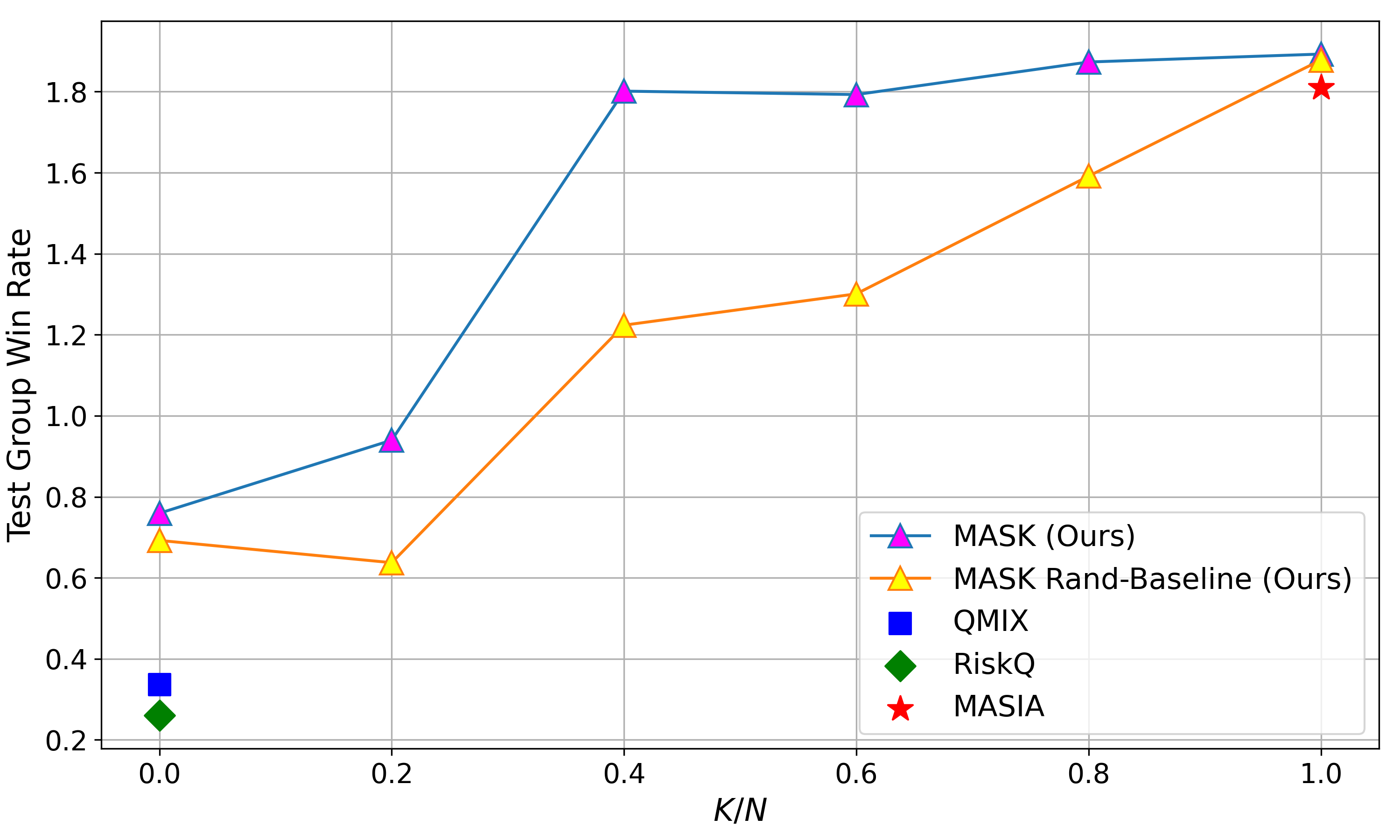}
    \caption{Test group win rate as a function of the communication parameter $K$ in the Hallway Group environment with $N=5$ agents. For \method{} and the random baseline, $K$ denotes the number of transmitting agents.}
    \label{fig:comm_rate_ablation}
\end{figure}

To analyze the trade-off between channel access and coordination capability, we evaluate agent performance under varying physical constraints, where the number of available channels $K$ ranges from $0$ (no communication) to $N=5$ (full broadcast). Figure~\ref{fig:comm_rate_ablation} illustrates the test group win rates averaged over the entire training process across different $K$ values in the Hallway Group environment. We observe that performance improves as the number of available channels increases, with the most significant gains occurring between $K=0$ and $K=2$. Notably, our method reaches a performance plateau at $K=2$, achieving a win rate comparable to the full-communication MASIA baseline. This demonstrates that \method{} effectively filters semantically high-value information, achieving near-optimal coordination while utilizing only 2 channels (40\% of the swarm size). To further verify that the A-SIG module effectively identifies task-critical information, we introduce \method{} $\text{Rand-Baseline}$ where $K$ transmitting agents are selected uniformly at random. The performance gap between A-SIG and random baseline confirms that the learned importance scores successfully filter task-critical information rather than simply benefiting from increased channel access.

\section{Conclusion}
\label{sec:conclusion}

This work presented \method{}, a unified framework for 6G-connected robotics that jointly addresses partial observability, safety-critical decision-making, and strict channel access constraints. By synergizing Arbiter-Assisted Semantic Information Gating and a self-supervised global state encoder within a risk-sensitive distributional reinforcement learning architecture, \method{} transforms communication from a passive overhead into an active, optimizable resource. Our results demonstrate that agents can learn to transmit only task-critical observations, maintaining high-performance coordination and safety even when channel access is constrained over 50\% or channels are unreliable. This capability is essential for deploying robust, risk-aware robotic swarms in realistic wireless environments where spectral efficiency and mission safety are paramount. Future work will focus on distributed scheduling, scaling the global encoder for massive swarms, and adapting the framework for highly dynamic, asymmetric 6G network topologies.
\bibliography{icc}
\bibliographystyle{IEEEtran}

\end{document}